\documentclass{IEEEcsmag}

\usepackage {hyperref } 
\hypersetup {colorlinks,urlcolor=blue,linkcolor=blue,citecolor=blue}

\expandafter\def\expandafter\UrlBreaks\expandafter{\UrlBreaks\do\/\do\*\do\-\do\~\do\'\do\"\do\-}
\usepackage{upmath,color}

\usepackage{booktabs}       
\usepackage{amsfonts}       
\usepackage{nicefrac}       
\usepackage{microtype}      
\usepackage{xcolor}         
\usepackage{multirow}
\usepackage[flushleft]{threeparttable} 
\usepackage{amsmath}
\usepackage{mathtools}
\usepackage{amssymb}
\usepackage{multirow}
\usepackage{ulem}

\usepackage{wrapfig}

\jvol{XX}
\jnum{XX}
\paper{8}
\jmonth{November}
\jname{IEEE Micro}
\jtitle{NetDistiller: Empowering Tiny Deep Learning
via In-Situ Distillation}
\pubyear{2023}

\begin{document}

\sptitle{Theme Article: tinyML}

\title{NetDistiller: Empowering Tiny Deep Learning via In-Situ Distillation}

\author{Shunyao Zhang}
\affil{Rice University, Houston, TX, 77005, USA}

\author{Yonggan Fu}
\affil{Georgia Institute of Technology, Atlanta, GA, 30332, USA}

\author{Shang Wu}
\affil{Rice University, Houston, TX, 77005, USA}

\author{Jyotikrishna Dass}
\affil{Rice University, Houston, TX, 77005, USA}

\author{Haoran You}
\affil{Georgia Institute of Technology, Atlanta, GA, 30332, USA}

\author{Yingyan (Celine) Lin}
\affil{Georgia Institute of Technology, Atlanta, GA, 30332, USA}

\markboth{tinyML}{tinyML}

\begin{abstract}
Boosting the task accuracy of tiny neural networks (TNNs) has become a fundamental challenge for enabling the deployments of TNNs on edge devices which are constrained by strict limitations in terms of memory, computation, bandwidth, and power supply. 
To this end, we propose a framework called NetDistiller to boost the achievable accuracy of TNNs by treating them as sub-networks of a weight-sharing teacher constructed by expanding the number of channels of the TNN. Specifically, the target TNN model is jointly trained with the weight-sharing teacher model via (1) gradient surgery to tackle the gradient conflicts between them and (2) uncertainty-aware distillation to mitigate the overfitting of the teacher model. Extensive experiments across diverse tasks validate NetDistiller's effectiveness in boosting TNNs' achievable accuracy over state-of-the-art methods. Our code is available at \url{https://github.com/GATECH-EIC/NetDistiller}.

\end{abstract}

\maketitle

\label{sec:intro}

\begin{figure*}[t!]
    \centering
    \vspace{-1em}
    \includegraphics[width=0.99\linewidth]{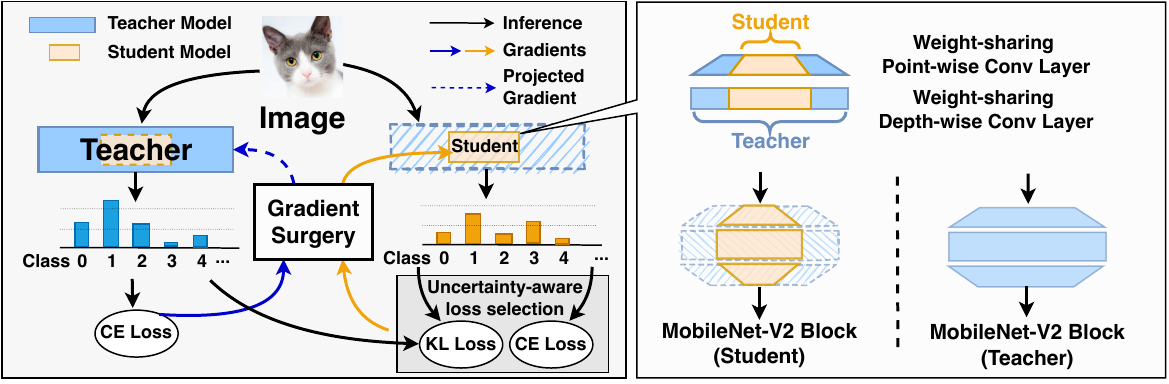}
    \vspace{-0.5em}

    \caption{An overview of NetDistiller. The target TNN is a  student model serving as a sub-network within a weight-sharing teacher model, constructed by expanding the number of channels of the target TNN. The teacher and student models are trained simultaneously while the teacher model is trained with the ground truth labels and the objective for training the student model is input-adaptively decided between an in-situ distillation mechanism and the ground truth labels based on its output uncertainty. To alleviate the gradient conflict issue observed during the training process, the teacher's gradients are modified via gradient surgery to remove the conflicting components based on the student's gradients.}
    \label{fig_overview}
    \vspace{-1.5em}
\end{figure*}

\chapteri{T}he recent record-breaking performance of neural networks (NNs) has spurred their increased application across various scientific and engineering disciplines. In parallel, 
it is projected that the worldwide number of Internet of Things (IoT) connected devices will reach 30.9 billion units by 2025~\cite{statista}. 
Deploying NN-powered intelligence on numerous IoT devices is of significant importance, as it enables the utilization of edge-collected data for various on-device intelligent functionalities that could potentially revolutionize human life. This tremendously growing demand has given rise to the field of tiny neural networks (TNNs) which have attracted substantially increasing attention. This is because TNNs enable small and inexpensive edge devices to work directly on local data at a lower power and computing cost, leading to both reduced latency and enhanced privacy as it alleviates or even eliminates the necessity of internet connectivity for sharing and centralizing the data on a cloud server. However, the achievable task performance of TNNs remains unsatisfactory due to their limited model capacity. Hence, improving the task performance of TNNs has become a fundamental challenge for enabling their wide-scale adoption, which is highly desired in numerous real-world edge applications.

To tackle the aforementioned challenge and thereby unlock the potential of TNNs at the edge, there has been an increasing research effort towards boosting their achievable task performance. In particular, it has been shown that training TNNs is fundamentally different from training large NNs. For example, the authors of \cite{cai2022network} identify that TNNs suffer from under-fitting due to their limited model capacity, in contrast to large NNs, which are prone to over-fitting. They also observe that data augmentation and regularization techniques, which enhance ImageNet accuracy for over-parameterized large NNs like ResNet50, have a detrimental effect on the accuracy of TNNs, such as MobileNetV2-Tiny \cite{lin2020mcunet}, which is 174$\times$ smaller than ResNet50.

Drawing inspiration from prior arts, we hypothesize that augmenting the model capacity (e.g., channels) during training enables TNNs to acquire additional knowledge, resulting in improved task accuracy. In vanilla knowledge distillation \cite{hinton2015distilling}, the knowledge encoded by a large model is transferred to a smaller one by training the small model with the outputs and/or activations of the large model. As a result, the small model is able to achieve a higher accuracy by mimicking the behaviors of the large model. In this work, we advocate a new in-situ knowledge distillation scheme, which is orthogonal to the vanilla one, further boosting the achievable task performance of TNNs. We make the following contributions:

\begin{itemize}
    \item We are the first to demonstrate that integrating a weight-sharing supernet with in-situ distillation can serve as an effective training recipe for boosting the achievable task performance of TNNs. Specifically, we propose a framework named NetDistiller, which incorporates the target TNN as a student model within a weight-sharing supernet that acts as a teacher model to boost the task performance of the trained TNNs without incurring any inference overhead.

    \item We identify that vanilla in-situ distillation can cause serious gradient conflicts between the supernet teacher and the sub-network student (i.e., the target TNN). Specifically, we find that up to $50\%$ of the weight gradients in 
 the student model have negative cosine similarities with those of the teacher model. This results in poor convergence when these gradients accumulate on their shared weights. Furthermore, vanilla in-situ distillation tends to induce overfitting in the teacher model, thus diminishing the effectiveness of our in-situ distillation. 

    \item To alleviate both the two issues identified above, NetDistiller proposes two solutions: (1) remove the conflicting gradients by projecting the teacher’s conflicting gradients onto the normal plane of the student's gradients, and (2) integrate an uncertainty-aware distillation to dynamically select the student loss function between the Kullback–Leibler divergence and the cross-entropy loss based on the certainty of the student model's output. These enhancements enable NetDistiller to unleash the promising effectiveness of in-situ distillation in more favorably training TNNs. 

   \item We perform extensive evaluations and ablation studies to validate the effectiveness of our NetDistiller framework for boosting the achievable accuracy of TNNs when compared to the state-of-the-art (SOTA) method. For instance, we observe a $2.3\%$ higher accuracy over NetAug \cite{cai2022network}, when training the MobileNet-V3-w0.35 model on the ImageNet dataset. We understand that NetDistiller has opened up a new perspective for boosting the achievable task performance of TNNs and enriched the field of knowledge distillation.

\end{itemize}

\section{RELATED WORK}
\label{sec:related_work}

\textbf{Efficient / Tiny Neural Networks.}
Significant progress has been made in designing efficient and mobile-friendly NNs. For example, MobileNets \cite{howard2017mobilenets} utilize Depthwise Separable Convolutions, which replace a standard convolution layer with a combination of depthwise convolution and pointwise convolution, demonstrating the potential for reducing computational costs while maintaining task accuracy.
In parallel to manually designed efficient networks and compression schemes, automated machine learning has been successfully used via neural architecture search~\cite{tan2019efficientnet,wu2019fbnet}.
In contrast to the above techniques, our proposed framework aims to improve the inference accuracy of TNNs via in-situ distillation where the target TNN architecture is used as a sub-network student of a weight-sharing supernet teacher model.

\textbf{Knowledge Distillation.} Knowledge Distillation (KD) 
\cite{NIPS2014_ea8fcd92} refers to the idea of transferring knowledge acquired by a pre-trained and over-parameterized (teacher) model to a small (student) model that is more suitable for edge deployment. Specifically, the small model usually has insufficient capacity to learn a concise knowledge representation, and KD empowers the student model to learn the exact behavior of the teacher model by mimicking the teacher's outputs at each level, i.e., soft labels. 
In this work, we advocate a new in-situ distillation scheme, namely, NetDistiller to enhance the task performance of TNNs. In contrast to vanilla KD where the teacher model is a large NN pre-trained on a different dataset and the student model is a separate smaller model, NetDistiller is an orthogonal approach that incorporates the TNN as a student sub-network within a weight-sharing supernet that acts as the teacher model for in-situ distillation.

\textbf{Network Augmentation.} The authors in \cite{cai2022network} propose network augmentation, known as NetAug, to boost the accuracy of tiny deep learning by alleviating the under-fitting issue. Specifically, NetAug dynamically augments the network during training, incorporating the tiny model as a sub-model within a larger model for auxiliary supervision in addition to its independent functionality. In contrast, NetDistiller provides an alternative scheme to boosting the performance of TNNs via in-situ knowledge distillation. 
In particular, the TNN in NetDistiller acts as a sub-network (student) in a static weight-sharing supernet (teacher) constructed by expanding the channels of the target TNN. 
In the experimental results section, we provide a comparative study of the proposed NetDistiller with the SOTA scheme NetAug and find that NetDistiller outperforms NetAug, e.g., achieves $2.3\%$ higher accuracy when training the MobileNet-V3-w0.35 model on the ImageNet dataset.

\section{The NetDistiller FRAMEWORK}
\label{sec:method}

Our proposed NetDistiller is a training recipe for boosting the accuracy of TNNs by incorporating the target TNN as a student model (sub-network) within a weight-sharing supernet, which acts as a teacher model. Through in-situ distillation, NetDistiller distills and transfers the knowledge from a supernet teacher model to the sub-network student model which is our target TNN. In this section, we first describe the construction of the weight-sharing supernet from the TNN followed by practical implementation of our in-situ distillation. Next, we describe techniques to resolve the gradient conflicts between the teacher and student models, as well as mitigate the over-fitting issue in the teacher model during the final training stage via uncertainty-aware distillation. 
Finally, we discuss the training and the inference overheads incurred by NetDistiller.

\subsection{NetDistiller's Enabler 0: Constructing the Weight-Sharing Teacher Model}
NetDistiller expands the target TNN's channels to construct a weight-sharing supernet that functions as the teacher model. Thus, the target TNN acts as a sub-network model. Both the student and teacher models share weights across all convolution layers while maintaining their respective Batch-Normalization layers, which account for different running statistics (i.e., means and variances) of their activation values. As a novel training recipe for augmenting the capacity of a target TNN to alleviate under-fitting issues and to boost accuracy, NetDistiller constructs a teacher model with $3\times$ channel numbers as the target TNN. Figure \ref{fig_overview} depicts the above-described construction of the weight-sharing teacher model from the target TNN.

\subsection{NetDistiller's Enabler 1: In-Situ Distillation}
The goal of our in-situ distillation is to stabilize the training of the supernet and improve the performance of sub-networks. In particular, TNNs are more likely to get stuck in local minimums due to insufficient capacity, which limits their performance compared to over-parameterized large NNs \cite{cai2022network}. To tackle this, NetDistiller integrates the target TNN as a sub-network student model within a weight-sharing supernet teacher model constructed by expanding the channels of the target NN as demonstrated in Figure \ref{fig_overview}. To the best of our knowledge, NetDistiller is the first to demonstrate that applying in-situ distillation to a weight-sharing supernet \cite{wang2021alphanet} can serve as an effective training recipe for boosting the achievable task performance of TNNs. 

Specifically, in-situ distillation leverages the `soft labels' predicted by the supernet as the supervision signals to the sub-network student model during each training iteration while using ground truth labels for the teacher model. Formally, at training iteration $n$, the supernet parameter $W$ is updated by
\[ W^{n} \leftarrow W^{n-1} + \eta g(W^{n-1}), \]
where $\eta$ is the learning rate, and:

\vspace{-0.5em}
\begin{equation}
\begin{split}
g(W^{n-1}) = \nabla_{W} \Big(& \mathcal{L_D}(W) + \\ 
& \mathcal{L}_{stu}\big([W, W_{stu}]; W^{n-1}\big) \Big) \Big\vert_{W=W^{n-1}}
\end{split}
\end{equation}
 
Here, $\mathcal{L_D}$ is the cross-entropy loss of the supernet teacher on a training dataset $\mathcal{D}$, $W$ and $W_{stu}$ denote the teacher and student models, respectively, and $\mathcal{L}_{stu}$ is the student loss modulated by uncertainty-aware distillation (introduced in the Enabler 3 section). 
Additionally, the distillation process in NetDistiller is single-shot, i.e., it is implemented in-situ during training without additional computation and memory cost, unlike two-step vanilla KD where a large model has to be trained first.

\subsection{NetDistiller's Enabler 2: Gradient Surgery for Resolving Gradient Conflicts}
\label{sec:pcgrad}
Considering that the gradients from both the student and the teacher models accumulate on the shared weights, we identify that vanilla in-situ distillation may cause serious gradient conflicts between the supernet teacher and sub-network student (target TNN). Specifically, we find that up to $50\%$ of the student model gradients have a negative cosine similarity with those of the teacher model. Inspired by the PCGrad \cite{yu2020gradient} which performs gradient surgery for multi-task learning, NetDistiller tackles this by projecting the conflicting teacher gradients to the normal plane of student gradients, thereby removing the conflicting components in the teacher gradients and improving the performance of TNNs. Specifically,
let $\nabla l_{stu}$ and $\nabla l_{tea}$ denote the gradients of the student and the teacher models, respectively. We define $\phi$ as the angle between the above two gradients and $g$ as the final gradient for updating the weights. In order to guarantee the student model training, we project the conflicting teacher's gradient, $proj(\nabla l_{tea})$, when the cosine similarity $cos (\phi) = \dfrac{\nabla l_{stu}.\nabla l_{tea}}{\Vert\nabla l_{stu}\Vert \Vert\nabla l_{tea}\Vert}$ is negative, which is formulated as follows:

\begin{small}
\begin{equation}
\begin{split}
g &= \nabla l_{stu} + proj(\nabla l_{tea}), \text{where} \\
 proj(\nabla l_{tea})&= \begin{cases}
\nabla l_{tea} - \dfrac{\nabla l_{tea}^T \nabla l_{stu}}{\|\nabla l_{stu}\|^2} \nabla l_{stu}, & \text{if} \cos (\phi) < 0 \\
\nabla l_{tea}, & \text{otherwise}
\end{cases}
\end{split}
\label{eq: PCGrad}
\end{equation}
\end{small}

\subsection{NetDistiller's Enabler 3: Uncertainty-aware Distillation}
\label{sec:uncertainty}
Since the student and the weight-sharing teacher models are jointly trained from scratch and the over-parameterized teacher model converges faster than the student one, we observe that the supernet teacher model suffers from over-fitting at the final training stage. 
In addition, \cite{narayan2022predicting} advocates that large models have the largest improvement on samples where the small model is most uncertain. As for certain examples, even those where the small model is not particularly accurate, large models are often unable to improve. Based on these insights, we hypothesize that the teacher model is not always a good teacher during the whole training. 

In light of this, we propose a technique called uncertainty-aware distillation (UD) to dynamically select the student loss functions between the Kullback–Leibler (KL) divergence and cross-entropy losses based on the certainty of the student model output (see Figure \ref{fig_overview}). Specifically, we measure the uncertainty via the entropy of the student outputs. When the entropy of the student output is high (i.e., uncertain), the student is distilled by the weight-sharing teacher via the KL divergence loss, otherwise, the student is trained by the ground truth label via the cross-entropy loss. We formulate this process as follows:

\vspace{-1em}
\begin{equation}
\mathcal{L}_{stu}= \begin{cases}
\texttt{KL(}W_{stu}(x), W(x)\texttt{)}, & \textit{uncertainty} \geq \text{T}  \\
\texttt{CE(}W_{stu}(x), y\texttt{)},  & \text{otherwise}
\end{cases}
\label{eq: uncertainty}
\end{equation}

Here, T denotes the uncertainty threshold; $\mathcal{L}_{stu}$ denotes the student model loss; \texttt{KL()} and \texttt{CE()} denote the KL divergence loss and the cross-entropy loss, respectively; $x$ and $y$ denote the input data and the ground truth labels; and \textit{uncertainty} denotes the entropy of the student model outputs $W_{stu}(x)$. 

\vspace{-1em}
\subsection{Analysis of Training and Inference Overhead}
In contrast to the two-step distillation process in vanilla KD, NetDistiller performs \textbf{\textit{one-shot}} in-situ distillation of knowledge from the supernet teacher to the sub-network student model without any additional computation and memory cost. Similar to that of NetAug, NetDistiller introduces \textbf{\textit{zero}} extra inference overhead because only the target TNN is used during inference, enabling the deployment of TNNs feasible on resource-constrained edge devices. Despite expanding the target TNN model by $3\times$, we observe a mere $20\%$ increase in the training time of NetDistiller to that of vanilla TNNs.

\begin{table*}[t]
\vspace{-0.5em}
\centering
\caption{Benchmark NetDistiller with SOTA methods for training TNNs. r160: The input image resolution is $160\times160$. w0.35: The model has $0.35\times$ number of channels than the vanilla one.}
\resizebox{0.99\textwidth}{!}{

\begin{tabular}{cccccccc} \toprule
\multicolumn{1}{c}{\multirow{2}{*}{Model}} & MobileNet-V2-Tiny & MCUNet & MobileNet-V3, r160 & \multicolumn{2}{c}{ProxylessNAS, r160} & \multicolumn{2}{c}{MobileNet-V2, r160} \\
\multicolumn{1}{c}{} & r144 & r176 & w0.35 & w0.35 & w1.0 & w0.35 & w1.0 \\ \midrule
Params & 0.75M & 0.74M & 2.2M & 1.8M & 4.1M & 1.7M & 3.5M \\
MACs & 23.5M & 81.8M & 19.6M & 35.7M & 164.1M & 30.9M & 154.1M \\ \midrule
Baseline & 51.7\% & 61.5\% & 58.1\% & 59.1\% & 71.2\% & 56.3\% & 69.7\% \\
NetAug \cite{cai2022network} & 53.3\% & 62.7\% & 60.3\% & 60.8\% & 71.9\% & 57.8\% & 70.6\% \\ 
In-situ & 54.1\% & 62.7\% & 62.1\% & 60.7\% & 71.2\% & 58.5\% & 71.2\% \\
In-situ + PCGrad \cite{yu2020gradient} & 54.5\% & 63.4\% & 62.3\% & 61.3\% & 72.5\% & 59.0\% & 72.0\% \\
\textbf{NetDistiller (ours)} & \textbf{54.8\%} & \textbf{64.2\%} & \textbf{62.6\%} & \textbf{61.5\%} & \textbf{72.8\%} & \textbf{59.3\%} & \textbf{72.6\%}
\\\bottomrule
\end{tabular}

}
\label{tab:benchmark}
\end{table*}

\section{EXPERIMENTAL RESULTS}
\label{sec:exp}

\subsection{Experiment Setup}


\begin{table}[t]
\centering
\caption{Ablation study of channel expansion rates on MobileNet-V2-w0.35 (MBV2-w0.35) and MobileNet-V3-w0.35 (MBV3-w0.35). Different teacher sizes in the first row indicate the channel expansion rates. Considering the limited improvement (0.2\%) between $\times 4$ and $\times 3$ teachers on MobileNet-V2-w0.35 model and the training efficiency, the teacher with $\times 3$ size is selected in NetDistiller.}
\resizebox{0.48\textwidth}{!}{
\begin{tabular}{cccccc}
\toprule
Teacher Size & Baseline & $\times 2$ & $\times 3$ & $\times 4$ & $\times 5$ \\
\midrule
MobileNet-V2-0.35 & 56.3\% & 58.0\% & 58.5\% & \textbf{58.7\%} & 58.3\% \\
MobileNet-V3-0.35 & 58.1\% & 61.3\% & \textbf{62.1\%} & 61.8\% & 61.8\% \\
\bottomrule
\end{tabular}
}
\label{tab:teacher_size}%
\end{table}

\textbf{Models}. We benchmark NetDistiller with SOTA TNNs training methods, e.g., NetAug \cite{cai2022network}, and KD~\cite{hinton2015distilling}, on five commonly adopted TNNs~\cite{lin2020mcunet,cai2022network}, including MobileNet-V2-Tiny, MobileNet-V2 (w0.35 and w1.0), MobileNet-V3, MCUNet (256kb-1mb), and ProxylessNAS (w0.35, w1.0). In particular, w0.35 indicates the models have 0.35 times of channels over the vanilla one (i.e., w1.0). Following the model definitions in \cite{cai2022network}, the channels of w0.35 models are round to products of 8.

\textbf{Datasets.} Following \cite{cai2022network}, we consider the ImageNet dataset with an input resolution of r144, r160, and r176 for different target TNNs. In the external knowledge distillation experiments, the input resolution for the external teacher is set to r224 to match its pre-training configuration. The object detection experiments are trained on the PASCAL VOC 2007+2012 datasets and evaluated on the PASCAL VOC 2007 validation set with an input resolution of r416.

\textbf{Evaluation Metrics.}
We evaluate NetDistiller and the baseline methods in terms of the top-1 accuracy on ImageNet and the average precision at IoU=$0.5$ (AP50) for the object detection on PASCAL VOC.

\textbf{Training Setting.} 
Following the training setting in \cite{cai2022network}, we train TNNs for $180$ epochs using an SGD optimizer with a momentum of $0.9$ and an initial learning rate of $0.4$  with a cosine learning rate scheduler. We adopt a learning rate warm-up for $5$ epochs and the gradients are clipped to $1.0$ during the whole training period. 
The label smoothing technique with a factor of $0.1$ is adopted when using the ground truth label. For the NetDistiller with uncertainty-aware distillation, we use the same training recipe but increase the training epochs to $360$. The uncertainty threshold T is set to $3.75$ based on the empirical observation of our ablation study.
All the ImageNet experiments are run on 8 GPUs with a batch size of $1024$. 
As a recent paper \cite{cai2022network} discovered that data augmentation and regularization could be harmful to TNN training, we only utilize standard data augmentations (e.g. random flip, random crop) and disable regularization methods like dropout and drop path.
For transfer learning on the object detection task, MobileNet-V2-w0.35 and MobileNet-V3-w0.35 models are connected with a YOLO-v4 head. All the object detection experiments are trained via an SGD optimizer with a momentum of 0.9 and an initial learning rate of 1e-4 decayed by a cosine learning rate scheduler for 100 epochs with a batch size of 8.

\begin{table}[t]
\centering
\caption{Ablation study of gradient surgery on MobileNet-V2-w0.35 and MobileNet-V3-w0.35 models for 360 epochs. We disable gradient surgery and calculate the cosine-similarity between the two gradients (teacher's and student's) of each convolutional layer. The percentage values shown under different epochs reflect the average ratio of the number of layers with negative cosine-similarity (gradient conflicts) w.r.t. the total number of layers in the model. }
\resizebox{0.48\textwidth}{!}{
\begin{tabular}{cccccc} \toprule
Epoch & 1 & 90 & 180 & 270 & 360 \\ \midrule
MobileNet-V2-w0.35 & 51.5\% & 40.1\% & 37.4\% & 39.4\% & 38.2\% \\
MobileNet-V3-w0.35 & 50.1\% & 45.2\% & 34.7\% & 38.5\% & 37.4\% \\
\bottomrule
\end{tabular}
}
\label{tab:gradient_conflict_ratio}%
\end{table}%

\begin{table}[t]
\centering
\vspace{-0.5em}
\caption{Ablation study of different uncertainty-aware distillation thresholds on MobileNet-V2-w0.35 and MobileNet-V2-w1.0. The first row is the thresholds. The uncertainty-aware distillation distills the student model if its output entropy (uncertainty) is higher than the threshold and trains the student model with ground truth labels otherwise.}
\resizebox{0.45\textwidth}{!}{
\begin{tabular}{cccc}\toprule
\multirow{2}{*}{Model} & \multicolumn{3}{c}{Uncertainty Threshold} \\ 
 & 2.5 & 3.75 & 5.0 \\\midrule
MobileNet-V2-w0.35 & 59.1\% & \textbf{59.3\%} & 58.9\% \\
MobileNet-V2-w1.0 & 71.9\% & \textbf{72.6\%} & 71.2\%
 \\ \bottomrule
\end{tabular}
}
\label{tab:uncertainty}%
\end{table}%

\begin{table*}[t]
\centering
\caption{Ablation studies of combinatining NetDistiller and External knowledge distillation. KD: Distill the target TNN with an external teacher (ImageNet pre-trained ResNet-50). NetDistiller w/o UD: Uncertainty-aware distillation is turned off in the external KD experiments. NetDistiller+KD: The external teacher distills both NetDistiller teacher and student models.}
\resizebox{0.95\textwidth}{!}{

\begin{tabular}{ccccc}
\toprule
Model & Baseline & KD & \multicolumn{1}{c}{NetDistiller w/o UD} & \multicolumn{1}{c}{NetDistiller+KD} \\ \midrule
MobileNet-V2-Tiny, r144 & 51.7\% & 53.7\% (+2.0\%) & 55.5\% (+3.8\%) & \textbf{56.1\% (+4.4\%)} \\
MobileNet-V2-w0.35, r160 & 56.3\% & 58.4\% (+2.1\%) & 59.0\% (+2.7\%) & \textbf{59.5\% (+3.2\%)} \\
MobileNet-V3-w0.35, r160 & 58.1\% & 61.6\% (+3.5\%) & 62.3\% (+4.2\%) & \textbf{62.5\% (+4.4\%)} \\
ProxylessNAS-w0.35, r160 & 59.1\% & 60.8\% (+1.7\%) & 61.3\% (+2.2\%) & \textbf{61.9\% (+2.8\%)}
\\ \bottomrule
\end{tabular}

}
\label{tab:external_kd}%
\end{table*}%

\subsection{Benchmark with SOTA TNN Training Methods}
\label{sec:benchmark}

As shown in Table \ref{tab:benchmark}, our observations are as follows:  (1) NetDistiller improves up to $4.5\%$ accuracy as compared to all baselines and gains $2.3\%$ higher accuracy than the most competitive baseline NetAug on MobileNet-V3-w0.35. (2) In-situ distillation contributes the most: TNNs trained solely through in-situ distillation can either match or surpass the SOTA TNNs training method, NetAug \cite{cai2022network}. (3) Among all the models, TNNs gain an accuracy boost of approximately $0.5\%$ after the introduction of PCGrad \cite{yu2020gradient} to mitigate gradient conflicts. (4) Our proposed NetDistiller, which combines the in-situ distillation with PCGrad and uncertainty-aware distillation, proves to be an effective training approach for enhancing the attainable task performance of TNNs.

\subsection{Ablation Studies of NetDistiller}
\label{sec:ablation}

\textbf{Channel Expansion Rates of the Teacher Model.}
Since NetDistiller expands the channels of TNNs to create a weight-sharing supernet as the teacher model, the channel expansion rate of the teacher model plays a crucial role in determining the effectiveness of the in-situ distillation mechanism. This is particularly important considering that a thin teacher model may have limited capacity, while an excessively wide teacher may not effectively transfer its information.
In order to identify the optimal channel expansion rate, we evaluate NetDistiller with $\times2$, $\times3$, $\times4$, and $\times5$ channel expansion rates on top of two TNNs, MobileNet-V2-w0.35 and MobileNet-V3-w0.35, respectively. As shown in Table \ref{tab:teacher_size}, we observe that: (1) all four teacher models enhance the accuracy of TNNs, demonstrating the overall efficacy of NetDistiller, and (2) MobileNet-V2-w0.35 and MobileNet-V3-w0.35 achieve the highest accuracy when the channel expansion rates are $\times4$ and $\times3$, respectively. To minimize the training overhead caused by the expanded teacher model, we by default set the channel expansion rate to $\times3$ in NetDistiller.

\textbf{Quantify the Gradient Conflicts.}
Because of the weight-sharing mechanism, the joint training of the student and teacher models leads to the accumulation of gradients on the same weights, inevitably resulting in gradient conflicts.
To verify the occurrence of gradient conflicts at different training stages, we calculate the ratio of layers with negative cosine similarity, averaged over the validation set, in relation to the total number of layers in the model throughout the training process. The results in Table \ref{tab:gradient_conflict_ratio} demonstrate that the teacher and student models indeed experience gradient conflicts, with as many as $51.5\%$ of layers exhibiting conflicting gradients. To address this issue, the adoption of gradient surgery leads to an accuracy improvement of $0.5\%$ and $0.2\%$ for MobileNet-V2-w0.35 and MobileNet-V3-w0.35, respectively, as shown in Table \ref{tab:benchmark}.

\begin{figure*}[t!]
    \centering
    \includegraphics[width=0.99\linewidth]{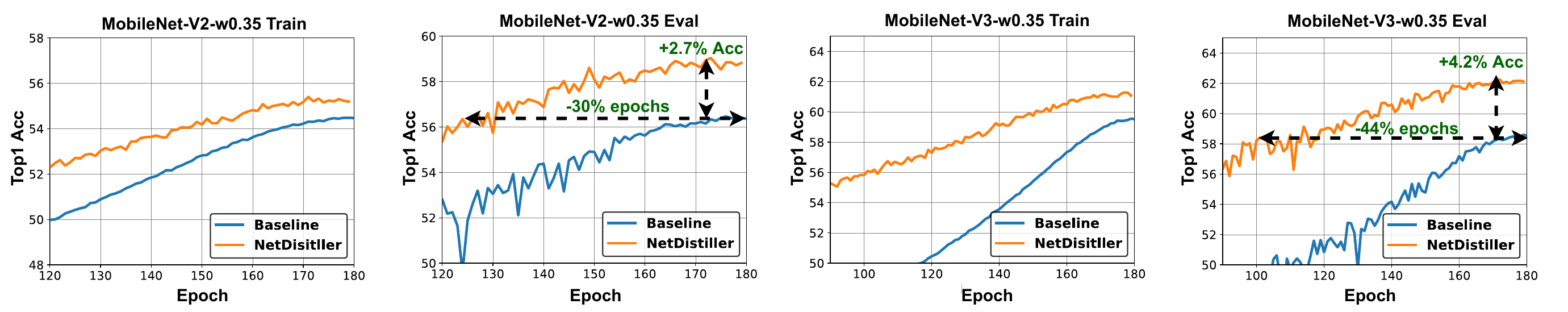}
    \caption{Visualizing the training process of NetDistiller and the baselines for MobileNet-V2-w0.35 and MobileNet-V3-w0.35 models. It reveals that NetDistiller significantly enhances both the training and evaluation accuracy of the TNNs. This highlights the substantial potential of NetDistiller in strengthening TNNs and mitigating the underfitting problem.}
    \label{fig: training_curve}
    \vspace{-1em}
\end{figure*}

\textbf{Threshold of the Uncertainty-aware Distillation.}
Our proposed uncertainty-aware distillation mechanism dynamically selects the objective for the student model, opting for either in-situ distillation or cross-entropy based on the uncertainty of the student model's outputs.
To decide the uncertainty threshold, we validate MobileNet-V2-w0.35 and MobileNet-V2-w1.0 models with uncertainty thresholds of $5.0$, $3.75$, and $2.5$, considering the entropy of ImageNet models is in the range of $[1.5, 10]$ when adopting a label smoothing factor of $0.1$. As shown in Table \ref{tab:uncertainty}, we observe that both models achieve their highest accuracy when the uncertainty threshold is set to $3.75$, resulting in accuracy improvements of $0.3\%$ and $0.6\%$, respectively, as compared to NetDistiller without uncertainty-aware distillation. Consequently, we adopt a default uncertainty threshold of $3.75$ when enabling uncertainty-aware distillation, without any additional overhead.

\textbf{Complementary Knowledge Distillation.}
A natural baseline for evaluating NetDistiller is standard knowledge distillation. Recent observations have suggested that significant gaps between the teacher and student models can lead to suboptimal knowledge distillation performance \cite{cho2019efficacy}. Therefore, we hypothesize that \underline{(a)} our proposed in-situ distillation serves as a more effective mechanism for enabling TNN training when compared to knowledge distillation, and \underline{(b)} knowledge distillation is complementary to our method and can be applied concurrently. To validate these hypotheses, we distill the knowledge from an ImageNet pre-trained ResNet-50 for both NetDistiller's teacher and student models. This process is referred to as external distillation to distinguish it from our in-situ distillation. We then benchmark this approach against (1) vanilla NetDistiller and (2) standard knowledge distillation. 
As shown in Table \ref{tab:external_kd}, we can observe that (1) vanilla NetDistiller can outperform standard knowledge distillation, e.g., an accuracy improvement of $3.8\%$ and $4.2\%$ on MobileNet-V2-Tiny and MobileNet-V3-w0.35 respectively, thus confirming our hypothesis (a); and (2) knowledge distillation is orthogonal to NetDistiller, as applying knowledge distillation on top of NetDistiller results in $4.4\%$ accuracy improvements on MobileNet-V2-Tiny and MobileNet-V3-w0.35, respectively, validating our hypothesis (b).

\subsection{Visualization of Training Trajectories}
We present the training curves for MobileNet-V2-w0.35 and MobileNet-V3-w0.35 throughout the 180 epochs of training using NetDistiller, as compared to the corresponding standard training baselines, in Figure \ref{fig: training_curve}. Notably, we observe that both the training and evaluation accuracy of NetDistiller consistently outperform the corresponding baselines at the same training epoch. For instance, NetDistiller achieves a $2.7\%$ accuracy improvement for MobileNet-V2-w0.35 and an impressive $4.2\%$ accuracy improvement for MobileNet-V3-w0.35 as compared to the baselines. Furthermore, to attain a comparable accuracy, NetDistiller requires fewer training epochs, resulting in a $44\%$ reduction in training epochs, as indicated in Figure \ref{fig: training_curve}.

\subsection{Transfer Learning Study on Object Detection}
To evaluate the generality of the representations learned by NetDistiller, we transfer NetDistiller's trained MobileNet-V2-w0.35 and MobileNet-V3-w0.35 to a downstream object detection task. We then compare their performance against standard pre-trained models, with or without knowledge distillation, on ImageNet.
Specifically, we replace the final pooling and linear layers in MobileNet-V2-w0.35 and MobileNet-V3-w0.35 with the YOLO-v4 object detection head.
As shown in Table \ref{tab:object detection}, NetDistiller consistently enjoys a better transferability with a higher Average Precision (AP) of $1.9\%$ / $1.6\%$  on MobileNet-V2-w0.35 / MobileNet-V3-w0.35 as compared to the standard training baselines. Note that although pre-trained features on classification tasks may not be necessarily useful for downstream tasks \cite{cai2022network}, which is also echoed with our results of KD pre-trained MobileNet-V3-w0.35 in Table \ref{tab:object detection}, NetDistiller manages to improve the achievable AP by up to $1.9\%$ on both models. This emphasizes the broad applicability of our method across diverse tasks and datasets.

\begin{table}[t]
\centering
\caption{Transfer learning on object detection tasks using PASCAL VOC 2007+2012 datasets and MobileNet-V2-w0.35 (MBV2) or MobileNet-V3-w0.35 (MBV3) models. Both of the two models are connected with the YOLO-v4 detection head. The model is initialized with NetDistiller pretrained weights on ImageNet. 
We report the Average Precision at IoU=$0.5$ (AP50). 
}

\resizebox{0.48\textwidth}{!}{
\begin{tabular}{cccc} \toprule
Model & Baseline AP50 & KD AP50 & NetDistiller AP50 \\ \midrule
MobileNet-v2-w0.35 & 60.4\% & 61.1\% & \textbf{62.3\%} \\
MobileNet-v3-w0.35 & 63.6\% & 62.8\% & \textbf{65.2\%} \\
\bottomrule
\end{tabular}
}
\label{tab:object detection}%
\end{table}%

\section{CONCLUSION}
\label{sec:conclusion}

Enhancing the task accuracy of TNNs is a critical challenge in enabling their deployment on resource-constrained IoT devices. Our proposed framework, NetDistiller, addresses this challenge by considering TNNs as sub-networks of a weight-sharing teacher model, achieved by expanding the number of channels in the TNN. By incorporating gradient surgery to handle gradient conflicts and uncertainty-aware distillation to alleviate teacher model overfitting, NetDistiller significantly improves the achievable accuracy of TNNs. Extensive experiments spanning various tasks demonstrate the superior effectiveness of NetDistiller compared to SOTA TNN training schemes. This advancement marks a significant step towards realizing the full potential of TNNs in real-world IoT applications.

\section{Acknowledgement}
\label{sec:ack}

The work is supported by the National Science Foundation (NSF) through the CCF program (Award number: 2211815) and by CoCoSys, one of the seven centers in JUMP 2.0, a Semiconductor Research Corporation (SRC) program sponsored by DARPA.



\def\refname{REFERENCES}


\bibliographystyle{unsrt}
\bibliography{ref}











\begin{IEEEbiography}{Shunyao Zhang}{\,} is a Ph.D. student at Rice University, Houston, USA. He received his master's degree in Electrical and Computer Engineering from Carnegie Mellon University, Pittsburgh, USA. His research interests are tiny ML and adversarial robustness. Contact him at sz74@rice.edu.
\end{IEEEbiography}

\begin{IEEEbiography}{Yonggan Fu}{\,} is a Ph.D. student at Georgia Institute of Technology. Before that, he obtained his Bachelor's degree from the School of The Gifted Young at the University of Science and Technology of China. His research focus and passion is to develop efficient and robust AI algorithms and co-design the corresponding hardware accelerators toward a triple-win in accuracy, efficiency, and robustness. Contact him at yfu314@gatech.edu.
\end{IEEEbiography}

\begin{IEEEbiography}{Shang Wu} {\,} is a master’s student at RICE University where he majored in Electrical and Computer Engineering. He received his bachelor’s degree in computer science at George Washington University. His research interests are in efficient ML, robust ML and generative AI. Contact him at sw99@rice.edu.
\end{IEEEbiography}

\begin{IEEEbiography}{Jyotikrishna Dass} {\,}is a Research Scientist at Rice University and manages the activities at Rice Data to Knowledge Program. His research interests are in distributed and parallel machine-learning systems for efficient edge computing. Previously, he was a postdoc at Dr. Yingyan Lin's lab. Dr. Dass received his Ph.D. in Computer Engineering from Texas A\&M University. Contact: jdass@rice.edu.
\end{IEEEbiography}

\begin{IEEEbiography}{Haoran You} {\,} is currently a Ph.D. student in the CS Department of Georgia Insitute of Technology. He received his bachelor's degree in the advanced class at Huazhong University of Science and Technology and his master's degree at Rice University. He is pursuing his doctoral degree in machine learning and computer architecture realm. His research interests include but are not limited to resource-constrained machine learning, computer vision, deep learning, and algorithm/accelerator co-design. Contact: haoran.you@gatech.edu.
\end{IEEEbiography}

\begin{IEEEbiography}{Yingyan (Celine) Lin} {\,} is currently an Associate Professor in the School of Computer Science and a member of the Machine Learning Center at the Georgia Institute of Technology. She leads the Efficient and Intelligent Computing (EIC) Lab. Her research focuses on developing efficient machine learning techniques via cross-layer innovations, spanning from efficient artificial intelligence (AI) algorithms to AI hardware accelerators and AI chip design, and aims to foster green AI and ubiquitous AI-powered intelligence. She received her Ph.D. degree in Electrical and Computer Engineering from the University of Illinois at Urbana-Champaign in 2017.
Prof. Lin has received the NSF CAREER Award, IBM Faculty Award, Facebook Faculty Research Award, and the ACM SIGDA Outstanding Young Faculty Award. She has served on the Technical Program committees for various conferences including DAC, ICCAD, MLSys, MICRO, and NeurIPS. She is currently an Associate Editor for the IEEE Transactions on Circuits and Systems II: Express Briefs. Contact her at celine.lin@gatech.edu.
\end{IEEEbiography}

\end{document}